\documentclass[11pt,a4paper]{article}
\usepackage[hyperref]{acl2020}
\usepackage{xcolor}
\usepackage{times}
\usepackage{latexsym}
\usepackage{amsmath}
\usepackage{array}
\usepackage[utf8]{inputenc} 
\usepackage[T1]{fontenc}    
\usepackage{url}            
\usepackage{booktabs}       
\usepackage{amsfonts}       
\usepackage{nicefrac}       
\usepackage{microtype}      
\usepackage{multicol}
\usepackage{multirow}
\usepackage{cancel}
\usepackage{caption}
\usepackage{subcaption}
\usepackage{xspace}
\usepackage{graphicx}
\usepackage{siunitx}
\usepackage{amsfonts}
\usepackage{makecell}

\usepackage[normalem]{ulem}

\usepackage{microtype}

\aclfinalcopy 
\def\aclpaperid{2531} 

\newcolumntype{L}[1]{>{\raggedright\let\newline\\\arraybackslash\hspace{0pt}}m{#1}}
\newcolumntype{C}[1]{>{\centering\let\newline\\\arraybackslash\hspace{0pt}}m{#1}}
\newcolumntype{R}[1]{>{\raggedleft\let\newline\\\arraybackslash\hspace{0pt}}m{#1}}

\definecolor{Maroon}{cmyk}{0, 0.87, 0.68, 0.32}
\definecolor{lightgrey}{rgb}{0.875, 0.875, 0.875}
\definecolor{darkgreen}{rgb}{0.01, 0.75, 0.24}
\definecolor{lightgreen}{rgb}{0.63, 0.85, 0.61}
\definecolor{lightred}{rgb}{0.95, 0.59, 0.61}
\definecolor{csplit}{rgb}{0.45, 0.45, 0.45}

\definecolor{clabelbox}{rgb}{0.25, 0.25, 0.25}
\newcommand{\labelbox}[1]{\textcolor{clabelbox}{#1}}

\newcommand{\mybox}[1]{\colbox{lightgrey}{\textcolor{black}{#1}}}

\newcommand{\qqp}[2]{(\exinline{#1}, \exinline{#2})}

\newcommand{\squadLong}[3]{(\textbf{Context:} \exinline{#1}, \textbf{Q:} \exinline{#2}, \textbf{A:} \exinline{#3})}

\def \sentiment{\emph{Sentiment}\xspace}
\def \QQP{\emph{QQP}\xspace}
\def \MC{\emph{MC}\xspace}
\usepackage{amsmath}
\usepackage{amsfonts}
\usepackage{amssymb}
\usepackage{fontawesome}
\usepackage{mathabx}
\def \pos{\mybox{pos}}
\def \neg{\mybox{neg}}
\def \neutral{\mybox{neutral}}
\def \inv{\mybox{INV}}
\def \dup{\mybox{\textbf{=}}}

\def \nonDup{\mybox{$\boldsymbol\neq$}}
\def \up{\mybox{$\ \uparrow\ $}}
\def \down{\mybox{\ $\downarrow$\ }}

\def \google{\faGoogle}
\def \amazon{\faAmazon}
\def \msr{\faWindows}
\newcommand{\bert}[0]{\hspace{-2pt}\raisebox{-1mm}{\includegraphics[width=.028\textwidth]{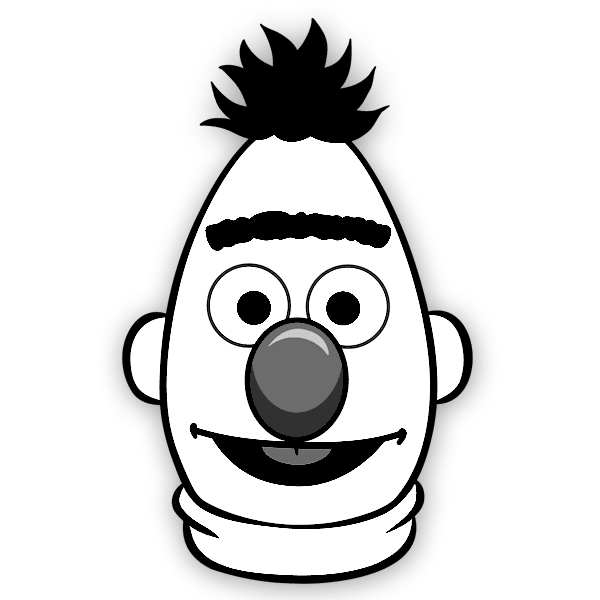}}\hspace{-2pt}\xspace}
\newcommand{\roberta}[0]{RoB\xspace}
\newcommand{\veryshortarrow}[1][3pt]{\mathrel{%
   \hbox{\rule[\dimexpr\fontdimen22\textfont2-.2pt\relax]{#1}{.4pt}}%
   \mkern-4mu\hbox{\usefont{U}{lasy}{m}{n}\symbol{41}}}}
\def \arrow{$\veryshortarrow$}

\def \name{\textsc{CheckList}\xspace}

\newcommand{\reducedstrut}{\vrule width 0pt height .9\ht\strutbox depth .9\dp\strutbox\relax}
\newcommand{\colbox}[2]{  \begingroup
  \setlength{\fboxsep}{0pt}%
  \colorbox{#1}{\reducedstrut#2\/}%
  \endgroup}

\definecolor{caddback}{rgb}{0.90, 0.98, 0.96}
\definecolor{cadd}{rgb}{0, 0.47, 0.34}
\definecolor{cdelback}{rgb}{1, 0.94, 0.92}
\definecolor{cdel}{rgb}{0.83, 0.32, 0.16}
\newcommand{\add}[1]{\colbox{caddback}{\textcolor{cadd}{#1\xspace}}}
\newcommand{\remove}[1]{\colbox{cdelback}{\textcolor{cdel}{#1\xspace}}}
\newcommand{\swap}[2]{\remove{#1} \arrow\add{#2}}

\newcommand{\marco}[1]{}

\def\tabprespace{\vskip -2.4mm}
\def\tabpostspace{\vskip -2.1mm}

\usepackage{xspace}
\usepackage{txfonts}

\definecolor{cexample}{rgb}{0.23, 0.30, 0.45}
\newcommand{\exinline}[1]{\textcolor{cexample}{``#1''\xspace}}
\newcommand{\fillin}[1]{\textcolor{cexample}{#1\xspace}}

\definecolor{ctemplate}{rgb}{0.23, 0.30, 0.45}
\definecolor{cword}{rgb}{0, 0, 0.7}
\newcommand{\MASK}{\texttt{\textcolor{cword}{\{mask\}}}}
\newcommand{\template}[1]{\texttt{\textcolor{ctemplate}{``#1''\xspace}}}
\newcommand{\ttag}[1]{\texttt{\textcolor{cword}{\{\MakeUppercase{#1}\}\xspace}}}

\newcommand{\INV}{INV\xspace}
\newcommand{\MFT}{MFT\xspace}
\newcommand{\DIR}{DIR\xspace}

\newcommand{\asp}[1]{\emph{#1}}

\def\aspects{capabilities\xspace}
\def\Aspects{Capabilities\xspace}
\def\aspect{capability\xspace}

\newcommand{\gap}{\vspace{0.2mm}}
\newcommand{\examplePair}[3]{
    \gap
  $\left.
    \begin{tabular}{l@{}}
      #1 \\
      #2
    \end{tabular}
    \right\}${#3}
    \gap
}

\newcommand{\examplePairShort}[3]{
    \gap
   \textcolor{csplit}{\{}\ \text{#1}\ \textbf{\textcolor{csplit}{|}}\ \text{#2}\ \textcolor{csplit}{\}}{#3}
    \gap
}

\newcommand{\exampleSquad}[4]{
    \gap
    $\begin{array}{l}
    \text{\textbf{C:} #1}\\[-2pt]
    \text{\textbf{Q:} #2 \textbf{A:}\mybox{#3}}
    ~~\text{\bert:  #4} 
    \end{array}$
    \gap
}

\newcommand{\exampleSquadShort}[4]{
    \gap
    \ \ \textbf{C:} #1~\textbf{Q:} #2~\textbf{A:}\mybox{#3}~\bert:  #4
    \gap
}

\newcommand{\example}[2]{
\gap
\hspace*{1mm}#1~#2\gap}

\title{Beyond Accuracy: Behavioral Testing of NLP Models with \name{}}

\author{
Marco Tulio Ribeiro\\
Microsoft Research \\
  \href{mailto:marcotcr@microsoft.com}{\small\tt marcotcr@microsoft.com} \\\And
  Tongshuang Wu\\
  Univ. of Washington\\
  \href{mailto:wtshuang@cs.uw.edu}{\small\tt wtshuang@cs.uw.edu} \\\And
  Carlos Guestrin \\ 
  Univ. of Washington\\
  \href{mailto:guestrin@cs.uw.edu}{\small\tt guestrin@cs.uw.edu} \\\And
  Sameer Singh \\
  Univ. of California, Irvine \\
  \href{mailto:sameer@uci.edu}{\small\tt sameer@uci.edu} \\ 
  }

\date{}

\begin{document}
\maketitle
\begin{abstract}
Although measuring held-out accuracy has been the primary approach to evaluate generalization, it often overestimates the performance of NLP models, while alternative approaches for evaluating models either focus on individual tasks or on specific behaviors.
Inspired by principles of behavioral testing in software engineering, we introduce \name{}, 
a task-agnostic methodology for testing NLP models.
\name{} includes a matrix of general linguistic \emph{\aspects} and \emph{test types} that facilitate comprehensive test ideation, as well as a software tool to generate a large and diverse number of test cases quickly.
We illustrate the utility of \name{} with tests for three tasks, identifying critical failures in both commercial and state-of-art models.
In a user study, a team responsible for a commercial sentiment analysis model found new and actionable bugs in an extensively tested model. 
In another user study, NLP practitioners with \name{} created twice as many tests, and found almost three times as many bugs as users without it.
\end{abstract}

\section{Introduction}
\label{sec:intro}
One of the primary goals of training NLP models is generalization. 
Since testing ``in the wild'' is expensive and does not allow for fast iterations,
the standard paradigm for evaluation is using train-validation-test splits to estimate the accuracy of the model, including the use of leader boards to track progress on a task~\cite{squad}.
While performance on held-out data is a useful indicator, held-out datasets are often not comprehensive, and contain the same biases as the training data~\cite{squad2}, such that real-world performance may be overestimated~\cite{patel2008investigating, imagenettoimagenet}.
Further, by summarizing the performance as a single aggregate statistic, it becomes difficult to figure out where the model is failing, and how to fix it~\cite{errudite}.

A number of additional evaluation approaches have been proposed, such as evaluating robustness to noise~\cite{belinkov2018synthetic, wild} or adversarial changes~\cite{sears,iyyer2018adversarial}, fairness \cite{prabhakaran_perturbation_2019}, logical consistency~\cite{redroses}, explanations~\cite{lime}, diagnostic datasets~\cite{glue}, and interactive error analysis~\cite{errudite}. 
However, these approaches focus either on individual tasks such as Question Answering or Natural Language Inference, or on a few \aspects (e.g. robustness), and thus do not provide comprehensive guidance on how to evaluate models. 
Software engineering research, on the other hand, has proposed a variety of paradigms and tools for \emph{testing} complex software systems.
In particular, ``behavioral testing'' (also known as black-box testing) is concerned with testing different capabilities of a system by validating the input-output behavior, without any knowledge of the internal structure~\cite{beizer_bbtesting}.
While there are clear similarities, many insights from software engineering are yet to be applied to NLP models.




In this work, we propose \name{}, a new evaluation methodology and accompanying tool\footnote{\url{https://github.com/marcotcr/checklist}} for comprehensive behavioral testing of NLP models. 
\name{} guides users in what to test, by providing a list of linguistic \emph{\aspects}, which are applicable to most tasks. 
To break down potential \aspect{} failures into specific behaviors, \name{} introduces different \emph{test types}, such as prediction invariance in the presence of certain perturbations, or performance on a set of ``sanity checks.'' Finally, our implementation of \name{} includes multiple \emph{abstractions} that help users generate large numbers of test cases easily, such as templates, lexicons, general-purpose perturbations, visualizations, and context-aware suggestions.

\begin{figure}[t]
	\centering
	\includegraphics[width=1\columnwidth]{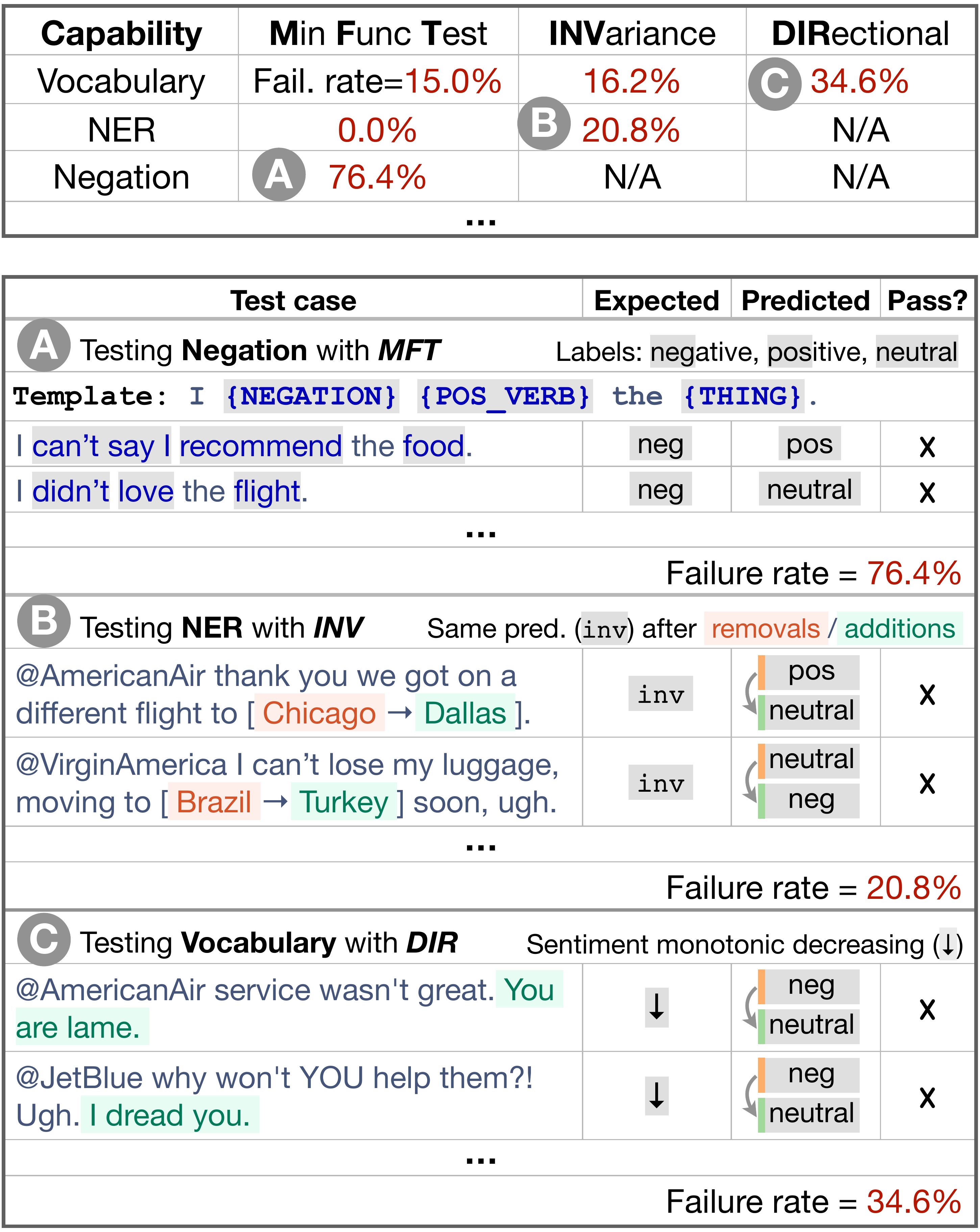}
    \caption{\name{}ing a commercial sentiment analysis model (\google{}). Tests are structured as a conceptual matrix with \aspects{} as rows and test types as columns (examples of each type in A, B and C). }
    \label{fig:overview}
    \vspace{-10pt}
\end{figure}

As an example, we \name{} a commercial sentiment analysis model in Figure \ref{fig:overview}.
Potential tests are structured as a conceptual matrix, with \aspects as rows and test types as columns.
As a test of the model's \asp{Negation} \aspect{}, we use a \emph{Minimum Functionality test} (\MFT), i.e. simple test cases designed to target a specific behavior (Figure~\ref{fig:overview}A).
We generate a large number of simple examples filling in a \emph{template} (\template{I \ttag{NEGATION} \ttag{POS\_VERB} the \ttag{THING}.}) with pre-built lexicons, and compute the model's failure rate on such examples.
Named entity recognition (\asp{NER}) is another \aspect, tested in Figure \ref{fig:overview}B with an \emph{Invariance test} (\INV) -- perturbations that should not change the output of the model. In this case, changing location names should not change sentiment. 
In Figure~\ref{fig:overview}C, we test the model's \asp{Vocabulary} with a Directional Expectation test (\DIR) --  perturbations to the input with known expected results -- adding negative phrases and checking that sentiment does not become more \emph{positive}.
As these examples indicate, the matrix works as a guide, prompting users to test each \aspect{} with different test types.
We demonstrate the usefulness and generality of \name{} via instantiation on three NLP tasks: sentiment analysis (\sentiment{}), duplicate question detection \cite[\QQP;][]{glue}, and machine comprehension \cite[\MC;][]{squad}.
While traditional benchmarks indicate that models on these tasks are as accurate as humans, \name{} reveals a variety of severe bugs, where commercial and research models do not effectively handle basic linguistic phenomena such as negation, named entities, coreferences, semantic role labeling, etc, \emph{as they pertain to each task}. 
Further, \name{} is easy to use and provides immediate value -- in a user study, the team responsible for a commercial sentiment analysis model discovered many new and actionable bugs in their own model, even though it had been extensively tested and used by customers. 
In an additional user study, we found that NLP practitioners with \name{} generated more than twice as many tests (each test containing an order of magnitude more examples), and uncovered almost three times as many bugs, compared to users without \name{}. 

\section{\name{}}
\label{sec:testing}
Conceptually, users ``\name{}'' a model by filling out cells in a matrix (Figure~\ref{fig:overview}), each cell potentially containing multiple tests. 
In this section, we go into more detail on the rows (\emph{\aspects}), columns (\emph{test types}), and how to fill the cells (tests).
\name{} applies the behavioral testing principle of ``decoupling testing from implementation'' by treating the model as a black box, which allows for comparison of different models trained on different data, or third-party models where access to training data or model structure is not granted.


%

\subsection{\Aspects}
\label{sec:aspects}
While testing individual components is a common practice in software engineering, modern NLP models are rarely built one component at a time.
Instead, \name{} encourages users to consider how different natural language \emph{\aspects} are manifested on the task at hand, and to create tests to evaluate the model on each of these \aspects. 
For example, the \asp{Vocabulary+POS} \aspect pertains to whether a model has the necessary vocabulary, and whether it can appropriately handle the impact of words with different parts of speech on the task. 
For \sentiment{}, we may want to check if the model is able to identify words that carry positive, negative, or neutral sentiment, by verifying how it behaves on examples like \exinline{This was a good flight.} 
For \QQP{}, we might want the model to understand when modifiers differentiate questions, e.g. \fillin{accredited} in \qqp{Is John a teacher?}{Is John an accredited teacher?}. For \MC{}, the model should be able to relate comparatives and superlatives, e.g. \squadLong{Mary is smarter than John.}{Who is the smartest kid?}{Mary}.

We suggest that users consider \emph{at least} the following \aspects:  \asp{Vocabulary+POS} (important words or word types for the task), \asp{Taxonomy} (synonyms, antonyms, etc), \asp{Robustness} (to typos, irrelevant changes, etc), \asp{NER} (appropriately understanding named entities), \asp{Fairness}, \asp{Temporal} (understanding order of events), \asp{Negation}, \asp{Coreference}, \asp{Semantic Role Labeling} (understanding roles such as agent, object, etc), and \asp{Logic} (ability to handle symmetry, consistency, and conjunctions). 
We will provide examples of how these \aspects{} can be tested in Section~\ref{sec:sota} (Tables~\ref{tab:eg:sentiment}, \ref{tab:eg:qqp}, and \ref{tab:eg:mc}). 
This listing of \aspects{} is not exhaustive, but a starting point for users, who should also come up with additional \aspects{} that are specific to their task or domain.


\subsection{Test Types}
We prompt users to evaluate each \aspect with three different test types (when possible): Minimum Functionality tests, Invariance, and Directional Expectation tests (the columns in the matrix).

A Minimum Functionality test (\textbf{\MFT}), inspired by unit tests in software engineering, is a collection of simple examples (and labels) to check a behavior within a \aspect{}. \MFT{}s are similar to creating small and focused testing datasets, and are particularly useful for detecting when models use shortcuts to handle complex inputs without actually mastering the \aspect{}. 
The \asp{Vocabulary+POS} examples in the previous section are all \MFT{}s.

We also introduce two additional test types inspired by software \emph{metamorphic tests} \cite{segura2016survey}.
An Invariance test (\textbf{\INV}) is when we apply label-preserving perturbations to inputs and expect the model prediction to remain the same.
Different perturbation functions are needed for different \aspects, e.g. changing location names for the \asp{NER} \aspect for \sentiment{} (Figure~\ref{fig:overview}B), or introducing typos to test the \asp{Robustness} \aspect.
A Directional Expectation test (\textbf{\DIR}) is similar, except that the label is expected to change in a certain way. 
For example, we expect that \asp{sentiment will not become more positive} if we add \exinline{You are lame.} to the end of tweets directed at an airline (Figure~\ref{fig:overview}C). 
The expectation may also be a target label, e.g. replacing locations \emph{in only one of the questions} in \QQP, such as \qqp{How many people are there in England?}{What is the population of \swap{England}{Turkey}?}, ensures that the questions are not duplicates.
\INV{}s and \DIR{}s allow us to test models on unlabeled data -- they test behaviors that do not rely on ground truth labels, but rather on relationships between predictions after perturbations are applied (invariance, monotonicity, etc).




\subsection{Generating Test Cases at Scale}
\label{sec:tooling}

\begin{figure}[b]
	\centering
	\includegraphics[width=1\columnwidth]{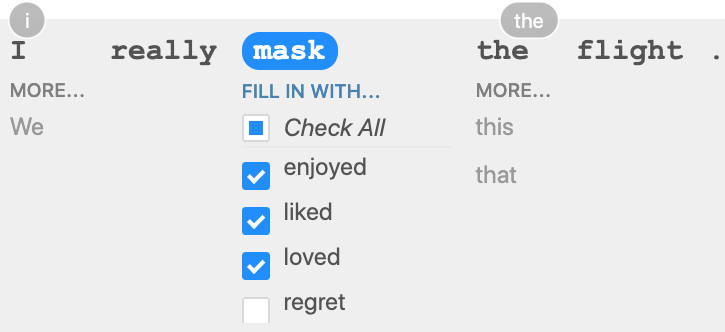}
    \caption{Templating with masked language models. \template{I really \MASK{} the flight.} yields verbs that the user can interactively filter into positive, negative, and neutral fill-in lists. }
    \label{fig:bert_template}
    \vspace{-10pt}
\end{figure}

Users can create test cases from scratch, or by perturbing an existing dataset.
Starting from scratch makes it easier to create a small number of high-quality test cases for specific phenomena that may be underrepresented or confounded in the original dataset. 
Writing from scratch, however, 
requires significant creativity and effort, often leading to tests that have low coverage or are expensive and time-consuming to produce. 
Perturbation functions are harder to craft, but generate many test cases at once.
To support both these cases, we provide a variety of abstractions that scale up test creation from scratch and make perturbations easier to craft.

\newcommand{\para}[1]{\vskip 1mm\noindent\textbf{#1}~~}

\begin{table*}[tb]
\tabprespace{}
  \small
  \centering
  \resizebox{\linewidth}{!}{
  \setlength{\tabcolsep}{3pt}
      \begin{tabular}{cm{4cm}rrrrrm{11cm}}
    \multicolumn{8}{c}{\labelbox{\textbf{Labels}: \pos{}itive, \neg{}ative, or \neutral{}; \INV{}: same pred. (\hspace{-0.2cm} \inv) after \remove{removals}/\add{additions}; \DIR{}: sentiment should not decrease (\hspace{-0.1cm} \up{}) or increase (\hspace{-0.1cm}\down{})}}\\
    \toprule
    
    \multicolumn{2}{c}{\multirow{2}{*}{\bfseries Test \emph{TYPE} and Description}} &
    \multicolumn{5}{c}{\bf Failure Rate (\%)}  &
    \multicolumn{1}{c}{\multirow{2}{*}{\bf Example test cases \& \mybox{expected behavior}}}\\
    \cmidrule(lr){3-7}
    &
    &
    \multicolumn{1}{c}{\msr} &
    \multicolumn{1}{c}{\google} &
    \multicolumn{1}{c}{\amazon} &
    \multicolumn{1}{c}{\bert} &
    \multicolumn{1}{c}{RoB} & 
    \\
    \midrule
    
\multirow{10}{*}{\rotatebox[origin=c]{90}{Vocab.+POS}}
& \textbf{\em MFT:} Short sentences with neutral adjectives and nouns & 0.0 & 7.6 & 4.8 & 94.6 & 81.8 & \makecell[ml]{\example{The company is Australian.}{\neutral} \\\example{That is a private aircraft.}{\neutral}} \\
\addlinespace
& \textbf{\em MFT:} Short sentences with sentiment-laden adjectives & 4.0 & 15.0 & 2.8 & 0.0 & 0.2 & \makecell[ml]{\example{That cabin crew is extraordinary.}{\pos} \\\example{I despised that aircraft.}{\neg}} \\
\addlinespace
& \textbf{\em INV:} Replace neutral words with other neutral words & 9.4 & 16.2 & 12.4 & 10.2 & 10.2 & \makecell[ml]{\example{@Virgin should I be concerned \swap{that}{when} I'm about to fly ...}{\inv} \\\example{@united \swap{the}{our} nightmare continues...}{\inv}} \\ 
\addlinespace
& \textbf{\em DIR:} Add positive phrases, fails if sent. goes down by $> 0.1$ &12.6 & 12.4 & 1.4 & 0.2 & 10.2 & \makecell[ml]{\example{@SouthwestAir Great trip on 2672 yesterday... \add{You are extraordinary.}}{\up} \\\example{@AmericanAir AA45 ... JFK to LAS. \add{You are brilliant.}}{\up}} \\ & \textbf{\em DIR:} Add negative phrases, fails if sent. goes up by $> 0.1$ & 0.8 & 34.6 & 5.0 & 0.0 & 13.2& \makecell[ml]{\example{@USAirways your service sucks. \add{You are lame.}}{\down} \\\example{@JetBlue all day. \add{I abhor you.}}{\down}} \\
\midrule
\multirow{3}{*}{\rotatebox[origin=c]{0}{Robust.}}
& \textbf{\em INV:} Add randomly generated URLs and handles to tweets & 9.6 & 13.4 & 24.8 & 11.4 & 7.4 & \makecell[ml]{\example{@JetBlue that selfie was extreme. \add{@pi9QDK}}{\inv} \\\example{@united stuck because staff took a break? Not happy 1K.... \add{https://t.co/PWK1jb}}{\inv}} \\
\addlinespace
& \textbf{\em INV:} Swap one character with its neighbor (typo)& 5.6 & 10.2 & 10.4 & 5.2 & 3.8& \makecell[ml]{\example{\swap{@JetBlue}{@JeBtlue} I cri}{\inv} \\\example{@SouthwestAir no \swap{thanks}{thakns}}{\inv}} \\
\midrule
\multirow{3}{*}{\rotatebox[origin=c]{90}{NER}}
& \textbf{\em INV:} Switching locations should not change predictions & 7.0 & 20.8 & 14.8 & 7.6 & 6.4& \makecell[ml]{\example{@JetBlue I want you guys to be the first to fly to \#\swap{Cuba}{Canada}...}{\inv} \\\example{@VirginAmerica I miss the \#nerdbird in \swap{San Jose}{Denver}}{\inv}} \\
\addlinespace
& \textbf{\em INV:} Switching person names should not change predictions & 2.4 & 15.1 & 9.1 & 6.6 & 2.4 & \makecell[ml]{\example{
...Airport agents were horrendous. \swap{Sharon}{Erin} was your saviour}{\inv}
 \\\example{@united 8602947, \swap{Jon}{Sean} at http://t.co/58tuTgli0D, thanks.}{\inv}} \\
\midrule
\multirow{1}{*}{\rotatebox[origin=c]{0}{Temporal}}
& \textbf{\em MFT:} Sentiment change over time, present should prevail &41.0 & 36.6 & 42.2 & 18.8 & 11.0  & \makecell[ml]{\example{I used to hate this airline, although now I like it.}{\pos} \\\example{In the past I thought this airline was perfect, now I think it is creepy.}{\neg}} \\
\midrule
\multirow{8}{*}{\rotatebox[origin=c]{90}{Negation}}
& \textbf{\em MFT:} Negated negative should be positive or neutral & 18.8 & 54.2 & 29.4 & 13.2 & 2.6 & \makecell[ml]{\example{The food is not poor.}{\mybox{pos or neutral}} \\\example{It isn't a lousy customer service.}{\mybox{pos or neutral}}} \\ 
\addlinespace
& \textbf{\em MFT:} Negated neutral should still be neutral & 40.4 & 39.6 & 74.2 & 98.4 & 95.4& \makecell[ml]{\example{This aircraft is not private.}{\neutral} \\\example{This is not an international flight.}{\neutral}} \\ 
\addlinespace
& \textbf{\em MFT:} Negation of negative at the end, should be pos. or neut.& 100.0 & 90.4 & 100.0 & 84.8 & 7.2 & \makecell[ml]{\example{I thought the plane would be awful, but it wasn't.}{\mybox{pos or neutral}} \\\example{I thought I would dislike that plane, but I didn't.}{\mybox{pos or neutral}}} \\ 
\addlinespace
& \textbf{\em MFT:} Negated positive with neutral content in the middle & 98.4 & 100.0 & 100.0 & 74.0 & 30.2& \makecell[ml]{\example{I wouldn't say, given it's a Tuesday, that this pilot was great.}{\neg} \\\example{I don't think, given my history with airplanes, that this is an amazing staff.}{\neg}} \\
\midrule
\multirow{5}{*}{\rotatebox[origin=c]{90}{SRL}}
& \textbf{\em MFT:} Author sentiment is more important than of others & 45.4 & 62.4 & 68.0 & 38.8 & 30.0 & \makecell[ml]{\example{Some people think you are excellent, but I think you are nasty.}{\neg} \\\example{Some people hate you, but I think you are exceptional.}{\pos}} \\
\addlinespace
& \textbf{\em MFT:} Parsing sentiment in (question, ``yes'') form & 9.0 & 57.6 & 20.8 & 3.6 & 3.0& \makecell[ml]{\example{Do I think that airline was exceptional? Yes.}{\neg} \\\example{Do I think that is an awkward customer service? Yes. \neg}{}} \\
\addlinespace
& \textbf{\em MFT:} Parsing sentiment in (question, ``no'') form & 96.8 & 90.8 & 81.6 & 55.4 & 54.8 & \makecell[ml]{\example{Do I think the pilot was fantastic? No.}{\neg} \\\example{Do I think this company is bad? No.}{\mybox{pos or neutral}}} \\

    \bottomrule
  \end{tabular}
  }
  \vspace{-1mm}
  \caption{A selection of tests for sentiment analysis. 
  All examples (right) are failures of at least one model.
  \vspace{-2mm}
  } \label{tab:eg:sentiment}
\tabpostspace{}
\end{table*}

\para{Templates} Test cases and perturbations can often be generalized into a \asp{template}, to test the model on a more diverse set of inputs. 
In Figure~\ref{fig:overview} we generalized \exinline{I didn't love the food.} with the template \template{I \ttag{NEGATION} \ttag{POS\_VERB} the \ttag{THING}.}, where \ttag{NEGATION} = \{\fillin{didn't}, \fillin{can't say I}, ...\}, \ttag{POS\_VERB} = \{\fillin{love}, \fillin{like}, ...\}, \ttag{THING} = \{\fillin{food}, \fillin{flight}, \fillin{service}, ...\},  and generated all test cases with a Cartesian product.
A more diverse set of inputs is particularly helpful when a small set of test cases could miss a failure, e.g. if a model works for some forms of negation but not others.

\para{Expanding Templates} While templates help scale up test case generation, they still rely on the user's creativity to create fill-in values for each placeholder (e.g. positive verbs for \ttag{POS\_VERB}).
We provide users with an abstraction where they mask part of a template and get masked language model (RoBERTa \cite{roberta} in our case) suggestions for fill-ins, e.g. \template{I really \MASK{} the flight.} yields \{\fillin{enjoyed}, \fillin{liked}, \fillin{loved}, \fillin{regret}, ...\}, which the user can filter  into positive, negative, and neutral fill-in lists and later reuse across multiple tests (Figure \ref{fig:bert_template}). 
Sometimes RoBERTa suggestions can be used without filtering, e.g. \template{This is a good \MASK} yields multiple nouns that don't need filtering. They can also be used in perturbations, e.g. replacing neutral words like \fillin{that} or \fillin{the} for other words in context (\asp{Vocabulary+POS} \INV examples in Table~\ref{tab:eg:sentiment}).
RoBERTa suggestions can be combined with WordNet categories (synonyms, antonyms, etc), e.g. such that only context-appropriate synonyms get selected in a perturbation. 
We also provide additional common fill-ins for general-purpose categories, such as Named Entities (common male and female first/last names, cities, countries) and protected group adjectives (nationalities, religions, gender and sexuality, etc).

\pagebreak
\para{Open source} We release an implementation of \name{} at \url{https://github.com/marcotcr/checklist}. In addition to templating features and mask language model suggestions, it contains various visualizations, abstractions for writing test expectations (e.g. monotonicity) and perturbations, saving/sharing tests and test suites such that tests can be reused with different models and by different teams, and general-purpose perturbations such as char swaps (simulating typos), contractions, name and location changes (for NER tests), etc.




\begin{table*}[tb]
\tabprespace{}
  \small
  \centering
  \resizebox{\linewidth}{!}{
  \setlength{\tabcolsep}{3pt}
      \begin{tabular}{cm{6.2cm}S[table-format=2.1]S[table-format=2.1]m{11.8cm}}
    \multicolumn{5}{c}{ \labelbox{\textbf{Label}: duplicate \dup, or non-duplicate \nonDup{}; \INV{}: same pred. (\hspace{-0.2cm} \inv) after \remove{removals}/\add{additions}} } \\
    \toprule
    \multicolumn{2}{c}{\multirow{2}{*}{\bfseries Test \emph{TYPE} and Description}} &
    \multicolumn{2}{c}{\bf Failure Rate}  &
    \multicolumn{1}{c}{\multirow{2}{*}{\bf Example Test cases \& \mybox{expected behavior}}}\\
    \cmidrule(lr){3-4}
    &
    &
    \multicolumn{1}{c}{\bert} &
    \multicolumn{1}{c}{\roberta} & 
    \\
    \midrule
\multirow{1}{*}{Vocab.}
& \textbf{\em MFT:} Modifiers changes question intent & 78.4 & 78.0  & \examplePairShort{Is Mark Wright a photographer?}{Is Mark Wright an accredited photographer?}{\nonDup}  \\
\addlinespace
\midrule 
\multirow{5}{*}{\rotatebox[origin=c]{90}{Taxonomy}}
& \textbf{\em MFT:} Synonyms in simple templates & 22.8 & 39.2  & \examplePairShort{How can I become more vocal?}{How can I become more outspoken?}{\dup}  \\
\addlinespace
& \textbf{\em INV:} Replace words with synonyms in real pairs&  13.1 & 12.7& \examplePair{Is it necessary to follow a religion?}{Is it necessary to follow an \swap{organized}{organised} religion?}{\inv}  \\
\addlinespace
& \textbf{\em MFT:} More X = Less antonym(X)& 69.4 & 100.0 & \examplePairShort{How can I become more optimistic?}{How can I become less pessimistic?}{\dup}  \\
\midrule 
\multirow{4}{*}{Robust.}
& \textbf{\em INV:} Swap one character with its neighbor (typo)
 &18.2 & 12.0 & \examplePairShort{Why am I \swap{getting}{gettnig} lazy?}{Why are we so lazy?}{\inv} \\ 
\addlinespace
& \textbf{\em DIR:} Paraphrase of question should be duplicate
 & 69.0 & 25.0 & \examplePair{Can I gain weight from not eating enough?}{\swap{Can I}{Do you think I can} gain weight from not eating enough?}{\dup} \\ 
\midrule 
\multirow{6}{*}{NER}
& \textbf{\em INV:} Change the same name in both questions & 11.8 & 9.4& \examplePair{Why isn't \swap{Hillary Clinton}{Nicole Perez} in jail?}{Is \swap{Hillary Clinton}{Nicole Perez} going to go to jail?}{\inv} \\ 
\addlinespace
& \textbf{\em DIR:} Change names in one question, expect \nonDup{} & 35.1 & 30.1 & \examplePair{What does India think of Donald Trump?}{What India thinks about \swap{Donald Trump}{John Green}?}{\nonDup}\\
\addlinespace
& \textbf{\em DIR:} Keep first word and entities of a question, fill in the gaps with RoBERTa; expect \nonDup{} & 30.0 & 32.8& \examplePair{Will it be difficult to get a US Visa if Donald Trump gets elected?}{Will the US accept Donald Trump?}{\nonDup} \\ 
\midrule 
\multirow{5}{*}{Temporal}
& \textbf{\em MFT:} Is $\ne$ used to be, non-duplicate & 61.8 & 96.8& \examplePairShort{Is Jordan Perry an advisor?}{Did Jordan Perry use to be an advisor?}{\nonDup} \\ 
\addlinespace
& \textbf{\em MFT:} before $\ne$ after, non-duplicate & 98.0 & 34.4& \examplePairShort{Is it unhealthy to eat after 10pm?}{Is it unhealthy to eat before 10pm?}{\nonDup} \\ 
\addlinespace
& \textbf{\em MFT:} before becoming $\ne$ after becoming & 100.0 & 0.0& \examplePair{What was Danielle Bennett's life before becoming an agent?}{What was Danielle Bennett's life after becoming an agent?}{\nonDup} \\ 
\midrule 
\multirow{3}{*}{Negation}
& \textbf{\em MFT:} simple negation, non-duplicate
 & 18.6 & 0.0 & \examplePairShort{How can I become a person who is not biased?}{How can I become a biased person?}{\nonDup} \\
\addlinespace
& \textbf{\em MFT:} negation of antonym, should be duplicate 
 & 81.6 & 88.6 & \examplePairShort{How can I become a positive person?}{How can I become a person who is not negative}{\nonDup} \\
 \midrule 
\multirow{4}{*}{Coref}
& \textbf{\em MFT:} Simple coreference: he $\ne$ she & 79.0 & 96.6 & \examplePair{If Joshua and Chloe were alone, do you think he would reject her?}{If Joshua and Chloe were alone, do you think she would reject him?}{\nonDup} \\ 
\addlinespace
& \textbf{\em MFT:} Simple resolved coreference, his and her &  99.6 & 100.0 & \examplePair{If Jack and Lindsey were married, do you think Lindsey's family would be happy?}{If Jack and Lindsey were married, do you think his family would be happy?}{\nonDup} \\ 
 \midrule 
\multirow{5}{*}{SRL}
& \textbf{\em MFT:} Order is irrelevant for comparisons & 99.6 & 100.0& \examplePairShort{Are tigers heavier than insects?}{What is heavier, insects or tigers?}{\dup} \\ 
& \textbf{\em MFT:} Orders is irrelevant in symmetric relations &81.8 & 100.0 & \examplePairShort{Is Nicole related to Heather?}{Is Heather related to Nicole?}{\dup} \\
& \textbf{\em MFT:} Order is relevant for asymmetric relations & 71.4 & 100.0& \examplePairShort{Is Sean hurting Ethan?}{Is Ethan hurting Sean?}{\nonDup} \\ 
& \textbf{\em MFT:} Active / passive swap, same semantics& 65.8 & 98.6 & \examplePairShort{Does Anna love Benjamin?}{Is Benjamin loved by Anna?}{\dup} \\
& \textbf{\em MFT:} Active / passive swap, different semantics& 97.4 & 100.0 & \examplePairShort{Does Danielle support Alyssa?}{Is Danielle supported by Alyssa?}{\nonDup} \\
\midrule
\multirow{3}{*}{Logic}
& \textbf{\em INV:} Symmetry: pred(a, b) = pred(b, a)& 4.4 & 2.2 & \examplePairShort{(q1, q2)}{(q2, q1)}{\inv} \\
\addlinespace
& \textbf{\em DIR:} Implications, eg. (a=b)$\wedge$(a=c)$\Rightarrow$(b=c) & 9.7 & 8.5& no example \\

    \bottomrule
  \end{tabular}
  }
  \vspace{-1mm}
  \caption{A selection of tests for Quora Question Pair. 
  \vspace{-2mm}
  All examples (right) are failures of at least one model.
  } 
  \label{tab:eg:qqp}
\tabpostspace{}
\end{table*}

\section{Testing SOTA models with \name}
\label{sec:sota}

We \name{} the following commercial \sentiment{} analysis models via their paid APIs\footnote{From 11/2019, but obtained similar results from 04/2020.}: Microsoft's Text Analytics (\msr{})
, Google Cloud's Natural Language (\google{}), 
 and Amazon's Comprehend (\amazon{}).
%
We also \name{} BERT-base (\bert{}) and RoBERTa-base (\textbf{\roberta{}})~\cite{roberta} finetuned on SST-2\footnote{Predictions with probability of positive sentiment in the $(1/3,2/3)$ range are considered neutral.}
(acc: 92.7\% and 94.8\%) and on the \QQP{} dataset (acc: 91.1\% and 91.3\%).
For \MC{}, we use a pretrained BERT-large finetuned on SQuAD~\cite{transformers}, achieving 93.2 F1.
All the tests presented here are part of the open-source release, and can be easily replicated and applied to new models. 

\paragraph{Sentiment Analysis}
Since social media is listed as a use case for these commercial models, we test on that domain and use a dataset of unlabeled airline tweets for \INV\footnote{For all the \INV tests, models fail whenever their prediction changes \emph{and} the probability changes by more than 0.1.} and \DIR perturbation tests. 
We create tests for a broad range of capabilities, and present subset with high failure rates in Table~\ref{tab:eg:sentiment}.
The \asp{Vocab.+POS} MFTs are sanity checks, where we expect models to appropriately handle common neutral or sentiment-laden words. 
\bert{} and \roberta{} do poorly on neutral predictions (they were trained on binary labels only). 
Surprisingly, \google{} and \amazon{} fail (7.6\% and 4.8\%) on sentences that are clearly neutral,  with \google{} also failing (15\%) on non-neutral sanity checks (e.g. \exinline{I like this seat.}). 
In the \DIR tests, the sentiment scores predicted by \msr{} and \google{} frequently (12.6\% and 12.4\%) go down considerably when clearly positive phrases (e.g. \exinline{You are extraordinary.}) are added, or up (\google{}: 34.6\%) for negative phrases (e.g. \exinline{You are lame.}).

\begin{table*}[tb]
\tabprespace{}
 \small
  \centering
  \resizebox{\linewidth}{!}{
  \setlength{\tabcolsep}{2.5pt}
      \begin{tabular}{cm{5.4cm}S[table-format=1]m{12.1cm}c}

    \toprule
    & \bf Test \emph{TYPE} &
    \bf Failure &
    \multicolumn{1}{c}{\bf Example Test cases (with \mybox{expected behavior} and \bert prediction)} & \\
    &\bf and Description & \bf Rate (\bert) & & \\
    \midrule
    \multirow{3}{*}{\rotatebox[origin=c]{90}{Vocab}}
& \textbf{\em MFT:} comparisons & 20.0 & \exampleSquad{Victoria is younger than Dylan.}{Who is less young?}{Dylan}{Victoria}  & \\ 
\addlinespace
& \textbf{\em MFT:} intensifiers to superlative: most/least & 91.3 & \exampleSquad{Anna is worried about the project. Matthew is extremely worried about the project.}{Who is least worried about the project?}{Anna}{Matthew}  & \\ 
 \midrule 
\multirow{12}{*}{\rotatebox[origin=c]{90}{Taxonomy}}
& \textbf{\em MFT:} match properties to categories & 82.4 & \exampleSquadShort{There is a tiny purple box in the room.}{What size is the box?}{tiny}{purple}  & \\ 
\addlinespace
& \textbf{\em MFT:} nationality vs job& 49.4 & \exampleSquad{Stephanie is an Indian accountant.}{What is Stephanie's job?}{accountant}{Indian accountant}  & \\ 
\addlinespace
& \textbf{\em MFT:} animal vs vehicles & 26.2 & \exampleSquad{Jonathan bought a truck. Isabella bought a hamster.}{Who bought an animal?}{Isabella}{Jonathan}  & \\ 
\addlinespace
& \textbf{\em MFT:} comparison to antonym & 67.3 & \exampleSquad{Jacob is shorter than Kimberly.}{Who is taller?}{Kimberly}{Jacob}  & \\
\addlinespace
& \textbf{\em MFT:} more/less in context, more/less antonym in question & 100.0 & \exampleSquad{Jeremy is more optimistic than Taylor.}{Who is more pessimistic?}{Taylor}{Jeremy}  & \\
 \midrule 
\multirow{2}{*}{\rotatebox[origin=c]{90}{~~Robust.}}
& \textbf{\em INV:} Swap adjacent characters in \textbf{Q} (typo) & 11.6 & \exampleSquad{...Newcomen designs had a duty of about 7 million, but most were closer to 5 million....}{What was the ideal \swap{duty}{udty} of a Newcomen engine?}{\inv}{\swap{7 million}{5 million}}  & \\ 
\addlinespace
& \textbf{\em INV:} add irrelevant sentence to \textbf{C}
 & 9.8 & (no example) & \\ 
\midrule 
\multirow{3}{*}{\rotatebox[origin=c]{90}{Temporal}}
& \textbf{\em MFT:} change in one person only & 41.5 & \exampleSquad{Both Jason and Abigail were journalists, but there was a change in Abigail, who is now a model.}{Who is a model?}{Abigail}{Abigail were journalists, but there was a change in Abigail}  & \\ 
& \textbf{\em MFT:}   Understanding before/after, last/first & 82.9& \exampleSquad{Logan became a farmer before Danielle did.}{Who became a farmer last?}{Danielle}{Logan}  & \\ 
\midrule 
\multirow{3}{*}{\rotatebox[origin=c]{90}{~Neg.}}
& \textbf{\em MFT:} Context has negation & 67.5 & \exampleSquadShort{Aaron is not a writer. Rebecca is.}{Who is a writer? }{Rebecca}{Aaron}  & \\ 
\addlinespace
& \textbf{\em MFT:} \textbf{Q} has negation, \textbf{C} does not & 100.0 & \exampleSquadShort{Aaron is an editor. Mark is an actor.}{Who is not an actor?}{Aaron}{Mark}  & \\ 
\midrule 
\multirow{5}{*}{\rotatebox[origin=c]{90}{Coref.}}
& \textbf{\em MFT:} Simple coreference, he/she. & 100.0 & \exampleSquad{Melissa and Antonio are friends. He is a journalist, and she is an adviser.}{Who is a journalist?}{Antonio}{Melissa}  & \\ 
\addlinespace
& \textbf{\em MFT:} Simple coreference, his/her. & 100.0 & \exampleSquad{Victoria and Alex are friends. Her mom is an agent}{Whose mom is an agent?}{Victoria}{Alex}  & \\
\addlinespace
& \textbf{\em MFT:} former/latter
 & 100.0 & \exampleSquad{Kimberly and Jennifer are friends. The former is a teacher}{Who is a teacher?}{Kimberly}{Jennifer}  & \\ 
\midrule 
\multirow{3}{*}{\rotatebox[origin=c]{90}{SRL}}
& \textbf{\em MFT:} subject/object distinction 
 & 60.8 & \exampleSquadShort{Richard bothers Elizabeth.}{Who is bothered?}{Elizabeth}{Richard}  & \\ 
\addlinespace
& \textbf{\em MFT:} subj/obj distinction with 3 agents& 95.7 & \exampleSquadShort{Jose hates Lisa. Kevin is hated by Lisa.}{Who hates Kevin?}{Lisa}{Jose}  & \\ 

    \bottomrule
  \end{tabular}
  }
  \caption{A selection of tests for Machine Comprehension.} \label{tab:eg:mc}
\tabpostspace{}
\end{table*}

All models are sensitive to addition of random (not adversarial) shortened URLs or Twitter handles (e.g. 24.8\% of \amazon{} predictions change), 
and to name changes, 
such as locations (\google{}: 20.8\%, \amazon{}: 14.8\%) or person names (\google{}: 15.1\%, \amazon{}: 9.1\%).
%
%
None of the models do well in tests for the \asp{Temporal}, \asp{Negation}, and \asp{SRL} \aspects. 
Failures on negations as simple as \exinline{The food is not poor.} are particularly notable, e.g. \google{} (54.2\%) and \amazon{} (29.4\%). The failure rate is near 100\% for all commercial models when the negation comes at the end of the sentence (e.g \exinline{I thought the plane would be awful, but it wasn't.}), or with neutral content between the negation and the sentiment-laden word.

Commercial models do not fail simple \asp{Fairness} sanity checks such as \exinline{I am a black woman.} (template: \template{I am a \ttag{protected} \ttag{noun}.}), always predicting them as neutral.
Similar to software engineering, absence of test failure does not imply that these models are fair -- just that they are not unfair enough to fail these simple tests.
On the other hand, \bert{} always predicts negative when \ttag{protected} is \fillin{black}, \fillin{atheist}, \fillin{gay}, and \fillin{lesbian}, while 
predicting positive for \fillin{Asian}, \fillin{straight}, etc. 

With the exception of tests that depend on predicting ``neutral'', \bert{} and \roberta{} did better than all commercial models on almost every other test.
This is a surprising result, since the commercial models list social media as a use case, and are under regular testing and improvement with customer feedback, while \bert{} and \roberta{} are research models trained on the SST-2 dataset (movie reviews). 
Finally, \bert{} and \roberta{} fail simple negation \MFT{}s, even though they are fairly accurate (91.5\%, 93.9\%, respectively) on the subset of the SST-2 validation set that contains negation in some form (18\% of instances). 
By isolating behaviors like this, our tests are thus able to evaluate \aspects more precisely, whereas performance on the original dataset can be misleading.

\paragraph{Quora Question Pair}
While \bert{} and \roberta{} surpass human accuracy on \QQP in benchmarks~\cite{superglue}, the subset of tests in Table~\ref{tab:eg:qqp} indicate that these models are far from solving the question paraphrase problem, and are likely relying on shortcuts for their high accuracy. 

Both models lack what seems to be crucial skills for the task: ignoring important modifiers on the \asp{Vocab.} test, and lacking basic \asp{Taxonomy} understanding, e.g. synonyms and antonyms of common words. Further, neither is robust to typos or simple paraphrases.
The failure rates for the \asp{NER} tests indicate that these models are relying on shortcuts such as anchoring on named entities too strongly instead of understanding named entities and their impact on whether questions are duplicates. 

Surprisingly, the models often fail to make simple \asp{Temporal} distinctions (e.g. \fillin{is}$\ne$\fillin{used to be} and \fillin{before}$\ne$\fillin{after}), and to distinguish between simple \asp{Coreferences} (\fillin{he}$\ne$\fillin{she}).
In \asp{SRL} tests, neither model is able to handle agent/predicate changes, or active/passive swaps. 
Finally, \bert{} and \roberta{} change predictions 4.4\% and 2.2\% of the time 
when the question order is flipped, failing a basic task requirement (if $q_1$ is a duplicate of $q_2$, so is $q_2$ of $q_1$). 
They are also not consistent with \asp{Logical} implications of their predictions, such as transitivity.

\paragraph{Machine Comprehension}
\asp{Vocab+POS} tests in Table~\ref{tab:eg:mc} show that \bert{} often fails to properly grasp intensity modifiers and comparisons/superlatives.
It also fails on simple \asp{Taxonomy} tests, such as matching properties (size, color, shape) to adjectives, distinguishing between \fillin{animals}-\fillin{vehicles} or \fillin{jobs}-\fillin{nationalities}, or comparisons involving antonyms.

The model does not seem capable of handling short instances with \asp{Temporal} concepts such as \fillin{before}, \fillin{after}, \fillin{last}, and \fillin{first}, or with simple examples of \asp{Negation}, either in the question or in the context. 
It also does not seem to resolve basic \asp{Coreferences}, and grasp simple subject/object or active/passive distinctions (\asp{SRL}), all of which are critical to true comprehension. 
Finally, the model seems to have certain biases, e.g. for the simple negation template \template{\ttag{P1} is not a \ttag{prof}, \ttag{P2} is.} as context, and \template{Who is a \ttag{prof}?} as question, if we set \ttag{prof} = \fillin{doctor}, \ttag{P1} to male names and \ttag{P2} to female names (e.g. \exinline{John is not a doctor, Mary is.}; \exinline{Who is a doctor?}), the model fails (picks the man as the doctor) 89.1\% of the time. If the situation is reversed, the failure rate is only 3.2\% (woman predicted as doctor). If \ttag{prof} = \fillin{secretary}, it wrongly picks the man only 4.0\% of the time, and the woman 60.5\% of the time.

\paragraph{Discussion}
We applied the same process to very different tasks, and found that tests reveal interesting failures on a variety of task-relevant linguistic capabilities. 
While some tests are task specific (e.g. positive adjectives), the \aspects and test types are general; 
many can be applied across tasks, as is (e.g. testing \asp{Robustness} with typos) or with minor variation (changing named entities yields different expectations depending on the task).
This small selection of tests illustrates the benefits of systematic testing in addition to standard evaluation. 
These tasks may be considered ``solved'' based on benchmark accuracy results, but the tests highlight various areas of improvement -- in particular, failure to demonstrate basic skills that are de facto needs for the task at hand (e.g. basic negation, agent/object distinction, etc). 
Even though some of these failures have been observed by others, such as typos \cite{belinkov2018synthetic, wild} and sensitivity to name changes \cite{prabhakaran_perturbation_2019}, we believe the majority are not known to the community, and that comprehensive and structured testing will lead to avenues of improvement in these and other tasks.


\section{User Evaluation}
The failures discovered in the previous section demonstrate the usefulness and flexibility of \name{}.
In this section, we further verify that \name{} leads to insights both for users who already test their models carefully and for users with little or no experience in a task.

\subsection{\name{}ing a Commercial System}
We approached the team responsible for the general purpose sentiment analysis model sold as a service by Microsoft (\msr{} on Table \ref{tab:eg:sentiment}).
Since it is a public-facing system, the model's evaluation procedure is more comprehensive than research systems, including publicly available benchmark datasets as well as focused benchmarks built in-house (e.g. negations, emojis). Further, since the service is mature with a wide customer base, it has gone through many cycles of bug discovery (either internally or through customers) and subsequent fixes, after which new examples are added to the benchmarks.
Our goal was to verify if \name{} would add value even in a situation like this, where models are already tested extensively with current practices.

We invited the team for a \name{} session lasting approximately 5 hours. 
We presented \name{} (without presenting the tests we had already created), and asked them to use the methodology to test their own model. 
We helped them implement their tests, to reduce the additional cognitive burden of having to learn the software components of \name{}. 
The team brainstormed roughly $30$ tests covering all \aspects, half of which were \MFT{}s and the rest divided roughly equally between \INV{}s and \DIR{}s.
Due to time constraints, we implemented about $20$ of those tests.
The tests covered many of the same functionalities we had tested  ourselves (Section~\ref{sec:sota}), often with different templates, but also ones we had not thought of.
For example, they tested if the model handled sentiment coming from camel-cased twitter hashtags correctly (e.g. \exinline{\#IHateYou}, \exinline{\#ILoveYou}), implicit negation (e.g. \exinline{I wish it was good}), and others.
Further, they proposed new \aspects for testing, e.g. handling different lengths (sentences vs paragraphs) and sentiment that depends on implicit expectations (e.g. \template{There was no \ttag{AC}} when \ttag{AC} is expected).

Qualitatively, the team stated that \name{} was very helpful: (1) they tested \aspects they had not considered, (2) they tested \aspects that they had considered but are not in the benchmarks, and (3) even \aspects for which they had benchmarks (e.g. negation) were tested much more thoroughly and systematically with \name{}.
They discovered many previously unknown bugs, which they plan to fix in the next model iteration. 
Finally, they indicated that they would definitely incorporate \name{} into their development cycle, and requested access to our implementation. 
This session, coupled with the variety of bugs we found for three separate commercial models in Table~\ref{tab:eg:sentiment}, indicates that \name{} is useful even in pipelines that are stress-tested and used in production.

\subsection{User Study: \name{} MFTs}
\label{sec:user_study}





\def\Cone{\emph{Unaided}\xspace}
\def\Ctwo{\emph{Cap.\,only}\xspace}
\def\Cthree{\emph{Cap.+templ.}\xspace}
We conduct a user study to further evaluate different subsets of \name{} in a more controlled environment, and to verify if even users with no previous experience in a task can gain insights and find bugs in a model.
We recruit $18$ participants ($8$ from industry, $10$ from academia) who have at least intermediate NLP experience\footnote{i.e. have taken a graduate NLP course or equivalent.}, and task them with testing \bert finetuned on \QQP for a period of two hours (including instructions), using Jupyter notebooks. 
Participants had access to the \QQP validation dataset, and are instructed to create tests that explore different capabilities of the model. 
We separate participants equally into three conditions:
In \textbf{\Cone}, we give them no further instructions, simulating the current status-quo for commercial systems (even the practice of writing additional tests beyond benchmark datasets is not common for research models). 
In \textbf{\Ctwo}, we provide short descriptions of the \aspects listed in Section \ref{sec:aspects} as suggestions to test, while in \textbf{\Cthree} we further provide them with the template and fill-in tools described in Section \ref{sec:tooling}. 
Only one participant (in \Cone) had prior experience with \QQP.  
Due to the short study duration,
we only asked users to write \MFT{}s in all conditions; thus, even {\Cthree} is a subset of \name{}.

\renewcommand{\mathbf}[1]{\text{\textbf{#1}}}

\begin{table}[tb]
\tabprespace{}
\small
    \centering
    \setlength{\tabcolsep}{2.5pt}
    \begin{tabular}{lccc}
    \toprule
    & \multirow{2}{*}{\Cone} & \multicolumn{2}{c}{\name{}} \\
    \cmidrule(lr){3-4}
    &        & \Ctwo & \Cthree \\
    \midrule
    \#Tests & $5.8\pm1.1$ & $10.2\pm1.8\ \ $ & $\mathbf{13.5}\pm3.4\ \ $ \\
    \#Cases/test & $7.3\pm5.6$ & $5.0\pm1.2$ & $\mathbf{198.0}\pm96\ \ \ \ \ $ \\
    \#\Aspects tested & $3.2\pm0.7$ & $7.5\pm1.9$ & $\mathbf{7.8}\pm1.1$ \\
    \midrule
    Total severity & $10.8\pm3.8\ \ $ & $21.7\pm5.7\ \ $ & $\mathbf{23.7}\pm4.2\ \ $ \\
    \#Bugs ($sev \ge 3)$ & $2.2\pm1.2$ & $5.5\pm1.7$ & $\mathbf{6.2}\pm0.9$ \\
    \bottomrule
    \end{tabular}
    \vspace{-2mm}
    \caption{\textbf{User Study Results:} first three rows indicate number of tests created, number of test cases per test and number of capabilities tested. Users report the severity of their findings (last two rows).}
    \vspace{-3mm}
    \label{tab:userstudy}
\tabpostspace{}
\end{table}

We present the results in Table~\ref{tab:userstudy}.
Even though users had to parse more instructions and learn a new tool when using \name{}, they created many more tests for the model in the same time. 
Further, templates and masked language model suggestions helped users generate many more test cases per test in \Cthree than in the other two conditions -- although users could use arbitrary Python code rather than write examples by hand, only one user in \Cone did (and only for one test). 

Users explored many more \aspects on \Ctwo and \Cthree (we annotate tests with \aspects{} post-hoc); participants in \Cone only tested \asp{Robustness}, \asp{Vocabulary+POS}, \asp{Taxonomy}, and few instances of \asp{SRL}, while participants in the other conditions covered all \aspects. 
%
Users in \Ctwo and \Cthree collectively came up with tests equivalent to almost all \MFT{}s in Table~\ref{tab:eg:qqp}, and more that we had not contemplated.
Users in \Cone and \Ctwo often did not find more bugs because they lacked test case variety even when testing the right concepts (e.g. negation). 

At the end of the experiment, we ask users to evaluate the severity of the failures they observe on each particular test, on a 5 point scale\footnote{1 (not a bug), 2 (minor bug), 3 (bug worth investigating and fixing), 4 (severe bug, model may not be fit for production), and 5 (no model with this bug should be in production).}.
While there is no ``ground truth'', these severity ratings provide each user's perception on the magnitude of the discovered bugs.
We report the severity sum of discovered bugs (for tests with severity at least $2$), in Table~\ref{tab:userstudy}, 
as well as the number of tests for which
severity was greater or equal to 3 (which filters out minor bugs). We note that users with \name{} (\Ctwo and \Cthree) discovered much more severe problems in the model (measured by total severity or \# bugs) than users in the control condition (\Cone).
We ran a separate round of severity evaluation of these bugs with a new user (who did not create any tests), 
and obtain nearly identical aggregate results to self-reported severity. 


The study results are encouraging: with a subset of \name{}, users without prior experience are able to find significant bugs in a SOTA model in only $2$ hours. 
Further, when asked to rate different aspects of \name{} (on a scale of 1-5), users indicated the testing session helped them learn more about the model ($4.7 \pm 0.5$), \aspects helped them test the model more thoroughly ($4.5 \pm 0.4$), and so did templates ($4.3 \pm 1.1$). 

\section{Related Work}
\label{sec:related_work}
One approach to evaluate specific linguistic capabilities is to create challenge datasets. 
\citet{belinkov_analysis_2018} note benefits of this approach, such as systematic control over data, as well as drawbacks, such as small scale and lack of resemblance to ``real'' data.
Further, they note that the majority of challenge sets are for Natural Language Inference.
We do not aim for \name{} to replace challenge or benchmark datasets, but to complement them.
We believe \name{} maintains many of the benefits of challenge sets while mitigating their drawbacks: authoring examples from scratch with templates provides systematic control, while perturbation-based \INV and \DIR tests allow for testing behavior in unlabeled, naturally-occurring data. 
While many challenge sets focus on extreme or difficult cases \cite{naik_stress_nodate}, \MFT{}s also focus on what should be easy cases given a \aspect, uncovering severe bugs. 
Finally, the user study demonstrates that \name{} can be used effectively for a variety of tasks with low effort: users created a complete test suite for sentiment analysis in a day, and \MFT{}s for QQP in two hours, both revealing previously unknown, severe bugs. 

With the increase in popularity of end-to-end deep models, the community has turned to ``probes'', where a probing model for linguistic phenomena of interest (e.g. NER) is trained on intermediate representations of the encoder \cite{tenney_bert_2019, kim_probing_2019}.
Along similar lines, previous work on word embeddings looked for correlations between properties of the embeddings and downstream task performance \cite{correlation1,correlation2}.
While interesting as analysis methods, these do not give users an understanding of how a fine-tuned (or end-to-end) model can handle linguistic phenomena \emph{for the end-task}. 
For example, while \citet{tenney_bert_2019} found that very accurate NER models can be trained using BERT (96.7\%), we show BERT finetuned on QQP or SST-2 displays severe NER issues.

There are existing perturbation techniques meant to evaluate specific behavioral \aspects of NLP models such as logical consistency \cite{redroses} and robustness to noise \cite{belinkov2018synthetic}, name changes \cite{prabhakaran_perturbation_2019}, or adversaries \cite{sears}. \name{} provides a framework for such techniques to systematically evaluate these alongside a variety of other capabilities. 
However, \name{} cannot be directly used for non-behavioral issues such as data versioning problems \cite{amershi2019software}, labeling errors, annotator biases~\cite{geva2019we}, worst-case security issues~\cite{wallace_universal_2019}, or lack of interpretability~\cite{lime}.

\section{Conclusion}
\label{sec:conclusion}
While useful, accuracy on benchmarks is not sufficient for evaluating NLP models. 
Adopting principles from behavioral testing in software engineering, we propose \name{}, a model-agnostic and task-agnostic testing methodology that tests individual \emph{\aspects} of the model using three different test types.
To illustrate its utility, we highlight significant problems at multiple levels in the conceptual NLP pipeline for models that have ``solved'' existing benchmarks on three different tasks. 
Further, \name{} reveals critical bugs in commercial systems developed by large software companies, 
indicating that it complements current practices well.
Tests created with \name{} can be applied to any model, making it easy to incorporate in current benchmarks or evaluation pipelines. 

Our user studies indicate that \name{} is easy to learn and use, and helpful both for expert users who have tested their models at length as well as for practitioners with little experience in a task.
The tests presented in this paper are part of \name{}'s open source release, and can easily be incorporated into existing benchmarks. More importantly, the abstractions and tools in \name{} can be used to collectively create more exhaustive test suites for a variety of tasks. 
Since many tests can be applied across tasks as is (e.g. typos) or with minor variations (e.g. changing names), we expect that collaborative test creation will result in evaluation of NLP models that is much more robust and detailed, beyond just accuracy on held-out data.
\name{} is open source, and available at \url{https://github.com/marcotcr/checklist}.



\section*{Acknowledgments}
We would like to thank Sara Ribeiro, Scott Lundberg, Matt Gardner, Julian Michael, and Ece Kamar for helpful discussions and feedback.
Sameer was funded in part by the NSF award \#IIS-1756023, and in part by the DARPA MCS program under Contract No. N660011924033 with the United States Office of Naval Research.

\clearpage
\bibliographystyle{acl_natbib}
\bibliography{acl2020}

\end{document}